\documentclass{article}
\usepackage{times}

\usepackage[final]{corl_2018}

\usepackage[numbers]{natbib}
\usepackage{multicol}

\usepackage{graphicx} 
\usepackage{fixltx2e}
\usepackage{tabularx}
\usepackage{stfloats}
\usepackage{caption}
\usepackage{epstopdf}
\usepackage{booktabs}
\usepackage{array}
\newcolumntype{P}[1]{>{\centering\arraybackslash}p{#1}}
\usepackage{color}
\usepackage{multirow}
\usepackage{balance}
\usepackage{mathptmx} 
\usepackage{times} 
\usepackage{amsmath} 
\usepackage{amssymb}  
\usepackage{comment}
\usepackage{gensymb}
\usepackage{interval}
\usepackage{subcaption}
\usepackage{hyperref}
\usepackage{siunitx}
\usepackage{algorithm}
\usepackage{wrapfig}
\definecolor{alexey}{rgb}{1.0,0.0,0.0}
\definecolor{rene}{rgb}{1.0,0.0,0.0}
\definecolor{davide}{rgb}{1.0,0.0,0.0}

\newcommand{\camready}[1]{{\color{black}{#1}}}

\newcommand{\darkgrayed}[1]{\textcolor{black}{#1}}
\makeatletter
\newcommand*\titleheader[1]{\gdef\@titleheader{#1}}
\AtBeginDocument{%
  \let\st@red@title\@title
  \def\@title{%
    \vskip-3em
    \bgroup\normalfont\large\centering\@titleheader\par\egroup
    \vskip1.5em\st@red@title}
}
\makeatother

\titleheader{\darkgrayed{This paper has been accepted for publication at the Conference on Robot Learning (CORL), Zurich, 2018.}}

\title{Deep Drone Racing: \\Learning Agile Flight in Dynamic Environments}

\author{%
   \hspace{-4mm}
   Elia Kaufmann$^{1}$\thanks{These two authors contributed equally. Correspondence to \{\texttt{ekaufmann, loquercio}\} \texttt{@ifi.uzh.ch}},  Antonio Loquercio$^{1}$\footnotemark[1],  Ren\'{e} Ranftl$^{2}$, \\
   \textbf{Alexey Dosovitskiy}$^{2}$, \textbf{Vladlen Koltun}$^{2}$, \textbf{Davide Scaramuzza}$^{1}$\\
   \vspace{-3mm}\\
   $^1$ Robotics and Perception Group\\
   Depts. Informatics and Neuroinformatics\\
   University of Zurich and ETH Zurich\\
   $^2$ Intel Labs\\
}

\begin{document}
\maketitle

\vspace{-10mm}

\begin{abstract}
Autonomous agile flight brings up fundamental challenges in robotics, such as coping with unreliable state estimation, reacting optimally to dynamically changing environments, and coupling perception and action in real time under severe resource constraints. In this paper, we consider these challenges in the context of autonomous, vision-based drone racing in dynamic environments. Our approach combines a convolutional neural network (CNN) with a state-of-the-art path-planning and control system. The CNN directly maps raw images into a robust representation in the form of a waypoint and desired speed. This information is then used by the planner to generate a short, minimum-jerk trajectory segment and corresponding motor commands to reach the desired goal. We demonstrate our method in autonomous agile flight scenarios, in which a vision-based quadrotor traverses drone-racing tracks with possibly moving gates. Our method
does not require any explicit map of the environment and runs fully onboard. We extensively test the precision and robustness of the approach in simulation and in the physical world. We also evaluate our method against state-of-the-art navigation approaches and professional human drone pilots.
\end{abstract}

\keywords{Drone Racing, Learning Agile Flight, Learning for Control.\\
\textbf{Supplementary video}: \url{http://youtu.be/8RILnqPxo1s}}

\begin{figure}[h]
\centering
\begin{subfigure}{.33\textwidth}
  \centering
  \includegraphics[width=0.98\linewidth]{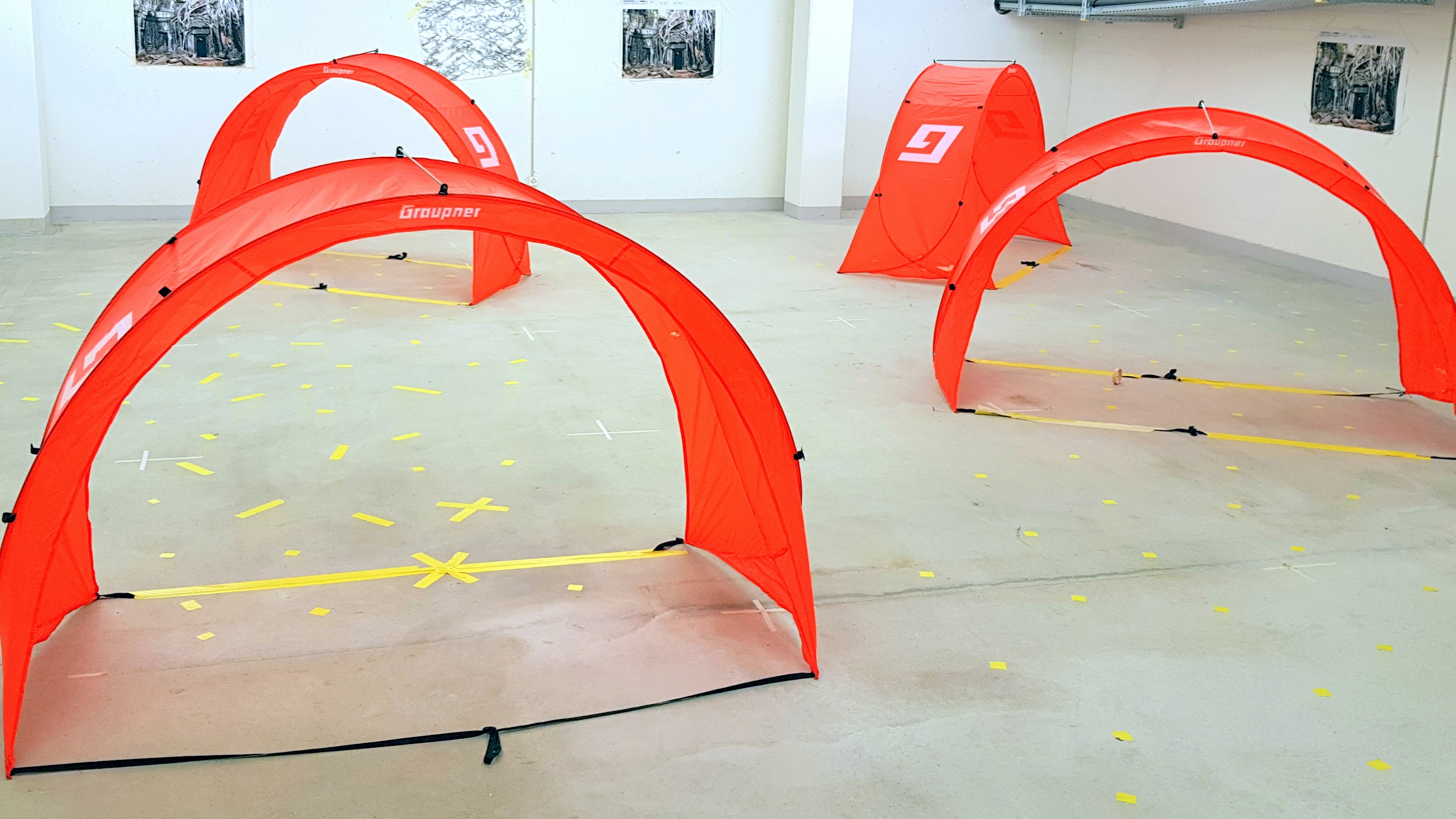}
  \caption{}
  \label{fig:catch_eye_1}
\end{subfigure}%
\begin{subfigure}{.33\textwidth}
  \centering
  \includegraphics[width=0.98\linewidth]{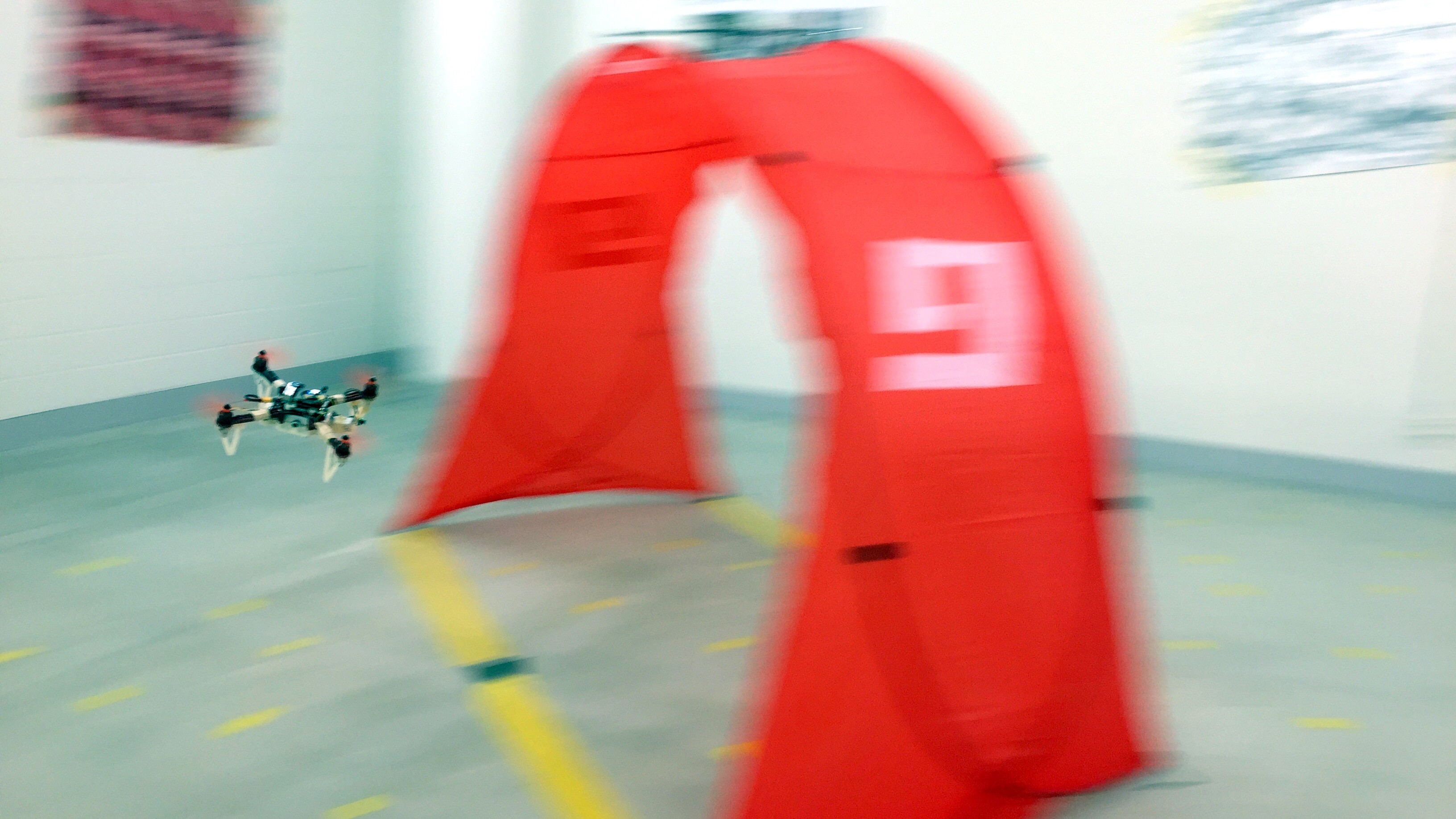}
  \caption{}
  \label{fig:catch_race}
\end{subfigure}
\begin{subfigure}{.33\textwidth}
  \centering
  \includegraphics[width=0.98\linewidth]{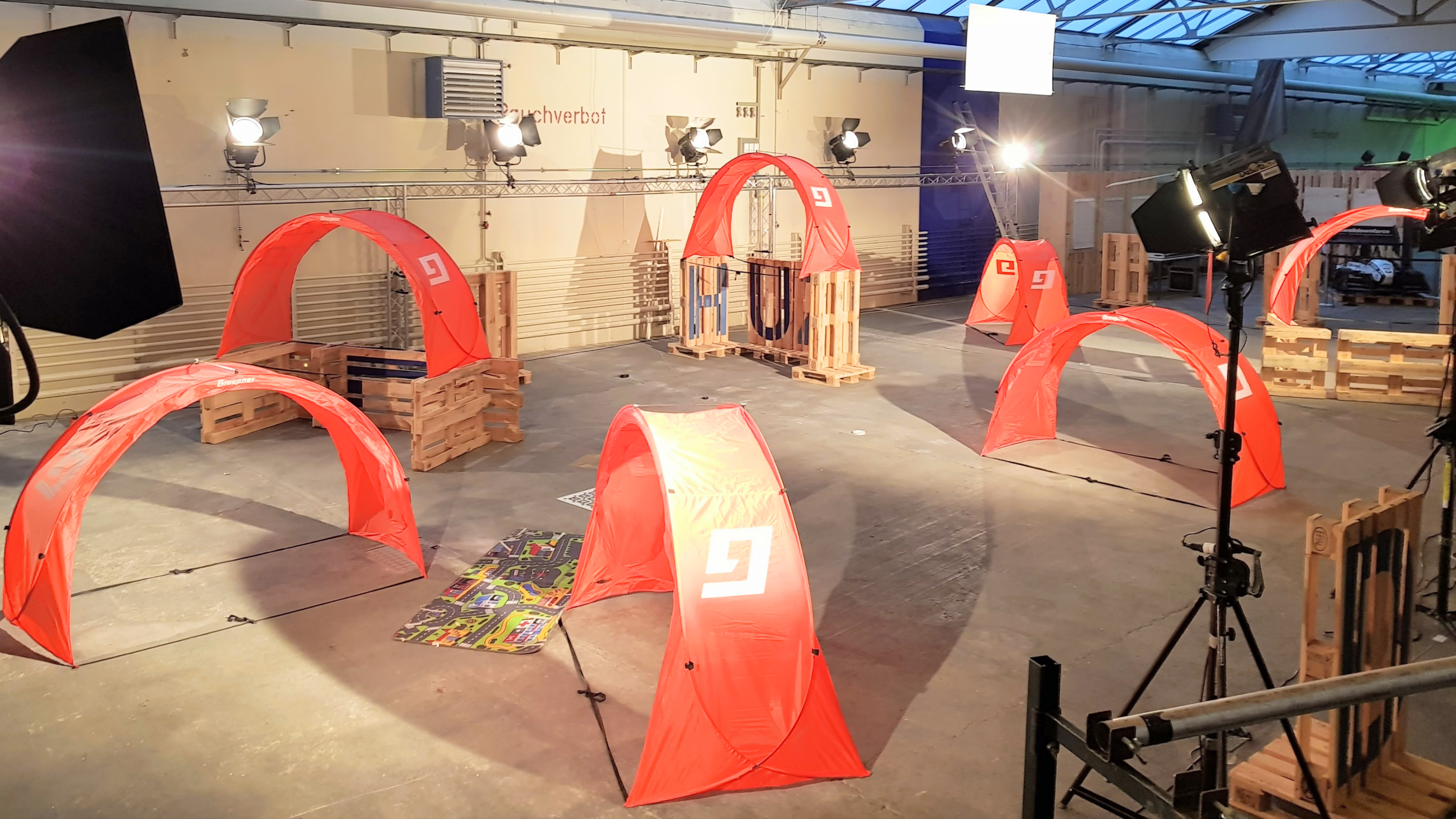}
  \caption{}
  \label{fig:catch_innopark}
\end{subfigure}
\caption{\small By combining a convolutional neural network with state-of-the-art trajectory generation and control methods, our vision-based, autonomous quadrotor is able to successfully navigate a race track with moving gates with high agility.}
	\label{fig:catch_eye}
	\vspace{-5mm}
\end{figure}

\section{Introduction}

Drone racing has become a popular televised sport with high-profile international competitions. In a drone race, each vehicle is controlled by a human pilot, who receives a first-person-view live stream from an onboard camera and flies the drone via a radio transmitter. Human drone pilots need years of training to master the advanced navigation and control skills that are required to be successful in international competitions. Such skills would also be valuable to autonomous systems that must quickly and safely fly through complex environments, in applications such as disaster response, aerial delivery, and inspection of complex structures.

%
%
We imagine that in the near future fully autonomous racing drones will compete against human pilots.
However, developing a fully autonomous racing drone is difficult due to challenges that span dynamics modeling, onboard perception, localization and mapping, trajectory generation, and optimal control.

A racing drone must complete a track in the shortest amount of time.
One way to approach this problem is to accurately track a precomputed global trajectory passing through all gates.
However, this requires highly accurate state estimation.
Simultaneous Localization and Mapping (SLAM) systems~\cite{Cadena2016} can provide accurate pose estimates against a previously-generated, globally-consistent map. These approaches may fail, however, when localizing against a map that was
created in significantly different conditions, or during periods of high acceleration (because of motion blur and loss of feature tracking). Additionally, enforcing global consistency leads to increased computational demands and significant difficulties in coping with dynamic environments. Indeed, SLAM methods enable navigation only in a predominantly-static world, where waypoints and (optionally) collision-free trajectories can be statically defined. In contrast, drone races (and related applications of flying robots) can include moving gates and dynamic obstacles.


In this paper, we take a step towards autonomous, vision-based drone racing in dynamic environments.
Our proposed approach is driven by the insight that methods relying on global state estimates in the form of robot poses are problematic due to the inherent difficulties of pose estimation at high speed along with inability to adequately cope with dynamic environments.
As an alternative, we propose a hybrid system that combines the perceptual awareness of a convolutional neural network (CNN) with the speed and accuracy of a state-of-the-art trajectory generation and tracking pipeline. Our method does not require an explicit map of the environment.
The CNN interprets the scene, extracts information from a raw image, and maps it to a robust representation in the form of a waypoint and desired speed.
This information is then used by the planning module to generate a short trajectory segment and corresponding motor commands to reach the desired local goal specified by the waypoint.
The resulting approach combines the advantages of both worlds: the perceptual awareness and simplicity of CNNs and the precision offered by state-of-the-art controllers.
The approach is both powerful and extremely lightweight: all computations run fully onboard.

Our experiments, performed in simulation and on a physical quadrotor, show that the proposed approach yields an integrated
perception and control system that is able to cope with highly dynamic environments and severe occlusions,
while being compact and efficient enough to be executed entirely onboard.
The presented approach can perform complex navigation tasks, such as seeking a moving gate or racing through a track, with higher robustness and precision than state-of-the-art, highly engineered systems.

\section{Related Work}
\label{sec:rel_work}
Pushing a robotic platform to high speeds poses a set of fundamental problems.
Motion blur, challenging lighting conditions, and perceptual aliasing can cause severe drifts in any state estimation system.
Additionally, state-of-the-art state estimation pipelines may require expensive sensors~\cite{lidar_slam}, have high computational costs~\cite{orb_slam}, or be subject to drift due to the use of compressed maps~\cite{get_out_lab}.
Therefore, real-time performance is generally hindered when operating with resource constrained platforms, such as small quadrotors.
This makes it extremely difficult to fully exploit the properties of popular minimum-snap or minimum-jerk trajectories~\cite{mellinger2011minimum, mueller2013computationally} for small, agile quadrotors using only onboard sensing and computing.
Moreover, state-of-the-art state estimation methods are designed for a \emph{predominantly-static world}, where no dynamic
changes to the environment, or to the track to follow, occur.

In order to cope with dynamic environments, it is necessary to develop methods that tightly couple the perception and action loops.
For the specific case of drone racing, this entails the capability to look for the target (the next gate) and localize relative to this while maintaining visual contact with it~\cite{aggressive_falanga, sayre2018visual}.
However, traditional, handcrafted gate detectors quickly become unreliable in the presence of occlusions, partial visibility, and motion blur.
The classical solution to this problem is visual servoing, where a robot is given a set of target locations in the form of reference images~\cite{classic_servo}.
However, visual servoing only works well when the difference between the current and the reference images is small
(which cannot be guaranteed at high speed) and is not robust to occlusions and motion blur.

An alternative solution consists of deriving actions directly from images using end-to-end trainable machine learning systems~\cite{safe_deep_nav, cad2rl, plato, deep_drive, Pan_2018, jung2018perception, Gupta_2017}.
While being independent of any global map and position estimate, these methods are not directly applicable to our specific problem due to their high computational complexity~\cite{cad2rl, Gupta_2017}, their low maximum speed~\cite{jung2018perception} or the inherent difficulties of generalizing to 3D motions~\cite{safe_deep_nav, deep_drive, Pan_2018}.
Furthermore, the optimal output representation for learning-based algorithms
that couple perception and control is an open question.
Known output representations range from predicting discrete navigation commands~\cite{Kim2015DeepNN,dronet}---
which enables high robustness but leads to low agility---to direct control~\cite{plato}---which can lead to highly agile control, but suffers from high sample complexity.

Taking the best of both worlds, this paper combines the benefits of agile trajectories with the ability of deep
neural networks to learn highly expressive perception models, which are able to reason on high-dimensional, raw sensory inputs.
The result is an algorithm that enables a resource-constrained, vision-based quadrotor to navigate a race track with possibly moving gates with high agility.
While the supervision to train our neural network comes from global trajectory methods,
the learned policy only operates on raw perceptual input, i.e., images, without requiring any information about the system's global state.
In addition, the ``learner'' acquires an ability that the ``teacher'' it imitates does not posses: it can cope with dynamic environments.

\section{Method}
\label{sec:methodology}

We address the problem of robust, agile flight of a quadrotor in a dynamic environment.
Our approach makes use of two subsystems: perception and control.
The perception system uses a Convolutional Neural Network (CNN) to predict a goal direction in local image coordinates, together with a desired navigation speed, from a single image from a forward-facing camera.
The control system then uses these outputs to generate a minimum jerk trajectory~\cite{mueller2013computationally} that is tracked by a low-level controller~\cite{Faessler15icra}.
In what follows we describe the subsystems in more detail.
%
\paragraph{Perception system}\label{sec:perception_system}
The goal of the perception system is to analyze the image and provide the flight direction to the control system.
We implement the perception system by a convolutional network.
The network takes as input a $300\times200$ RGB image, captured from the onboard camera, and outputs a tuple~$\lbrace \vec{x}, v \rbrace$, where $\vec{x} \in [-1,1]^2$ is a two-dimensional vector that encodes the direction to the new goal in normalized image coordinates and $v \in [0,1]$ is a normalized desired speed to approach it.
To allow for onboard computing, we employ the efficient ResNet-based architecture of Loquercio et al.~\cite{dronet} (see the supplement for details).
With our hardware setup, the network can process roughly $10$ frames per second onboard.
The system is trained by imitating an automatically computed expert policy, as explained in Section~\ref{sec:training_procedure}.
%

\paragraph{Control system}\label{sec:control_system}
Given the tuple $\lbrace \vec{x}, v \rbrace$, the control system generates low-level control commands.
To convert the goal position $\vec{x}$ from two-dimensional normalized image coordinates to three-dimensional local frame coordinates,
we back-project the image coordinates $\vec{x}$ along the camera projection ray and derive the
goal point at a depth equal to the prediction horizon $d$ (see Figure~\ref{fig:traj_sketch}).
%
%
We found setting $d$ proportional to the normalized platform speed $v$ predicted by the network to work well.
%
%
The desired quadrotor speed $v_{des}$ is computed by rescaling the predicted normalized speed $v$ by a user-specified maximum speed $v_{max}$: $v_{des} = v_{max} \cdot v$.
This way, with a single trained network, the user can control the aggressiveness of flight by varying the maximum speed.
Once $p_g$ in the quadrotor's body frame and $v_{des}$ are available,
a state interception trajectory $t_s$ is computed to reach the goal position (see Figure~\ref{fig:traj_sketch}).
Since we run all computations onboard, we use computationally efficient minimum jerk trajectories~\cite{mueller2013computationally} to generate $t_s$.
%
To track $t_s$, i.e., to compute the low-level control commands, we deploy the control scheme proposed by Faessler et al.~\cite{Faessler15icra}.

\subsection{Training procedure}\label{sec:training_procedure}
\begin{wrapfigure}[10]{r}[10pt]{0.5\linewidth}
\centering
\includegraphics[width=1\linewidth]{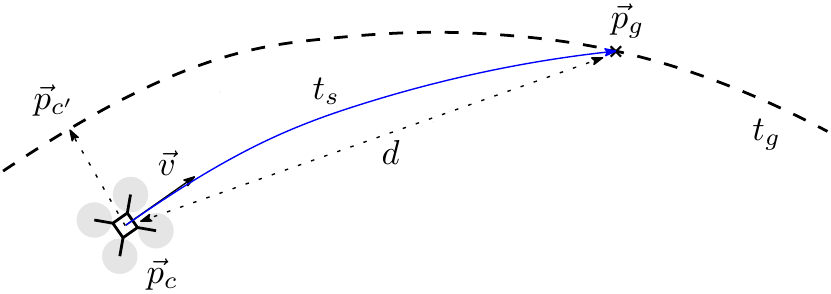}
\caption{Notation used for the control system. 
}
\label{fig:traj_sketch}
\end{wrapfigure}

We train the perception system with imitation learning, using automatically generated globally optimal trajectories as a source of supervision.
To generate these trajectories, we make the assumption that at training time the location of each gate of the race track, expressed in a common reference frame, is known.
Additionally, we assume that at training time the quadrotor has access to accurate state estimates with respect to this reference frame.
Note however that at test time no privileged information is needed and the quadrotor relies on image data only.
The overall training setup is illustrated in Figure~\ref{fig:traj_sketch}.

\textbf{Expert policy.}
We first compute a global trajectory~$t_{g}$ that passes through all gates of the track, using the minimum-snap trajectory implementation from~\citet{mellinger2011minimum}.
To generate training data for the perception network, we implement an expert policy that follows the reference trajectory.
Given a quadrotor position $\vec{p}_c \in \mathbb{R}^3$, we compute the closest point~$\vec{p}_{c'} \in \mathbb{R}^3$ on the global reference trajectory.
The desired position $\vec{p}_{g} \in \mathbb{R}^3$ is defined as the point on the global reference trajectory,
whose distance from $\vec{p}_c$ is equal to the prediction horizon $d \in \mathbb{R}$.
We project the desired position $\vec{p}_{g}$ onto the image plane of the forward facing camera to generate the ground truth normalized image coordinates $\vec{x}_{g}$ corresponding to the goal direction.
The desired speed $v_{g}$ is defined as the speed of the reference trajectory at~$\vec{p}_{c'}$ normalized by the maximum speed achieved along $t_g$.

\textbf{Data collection.}
To train the network, we collect a dataset of state estimates and corresponding camera images.
Using the global reference trajectory, we evaluate the expert policy on each of these samples and use the result as the ground truth for training.
An important property of this training procedure is that it is agnostic to how exactly the training dataset is collected.
The network is not directly imitating the demonstrated behavior, and therefore the performance of the learned policy is not upper-bounded by the performance of the provided demonstrations.

We use this flexibility to select the most suitable data collection method when training in simulation and in the real world.
The key consideration here is how to deal with the domain shift between training and test time.
(In our scenario, this domain shift mainly manifest itself when the quadrotor flies far from the reference trajectory $t_g$.)
In simulation, we employed a variant of DAgger~\cite{dagger}, which uses the expert policy to recover whenever the learned policy deviates far from the reference trajectory.
Repeating the same procedure in the real world would be infeasible: allowing a partially trained network to control a UAV would pose a high risk of crashing and breaking the platform.
Instead, we manually carried the quadrotor through the track and ensured a sufficient coverage of off-trajectory positions.

\textbf{Loss function.} We train the network with a weighted MSE loss on point and velocity predictions:
\begin{equation}
	L = \Vert \vec{x}  - \vec{x}_{g}\Vert^2 + \gamma (v - v_{g})^2,
	\label{eq:loss}
\end{equation}
where $\vec{x}_{g}$ denotes the groundtruth image coordinates and $v_{g}$ denotes the groundtruth speed.
By cross-validation, we found the optimal weight to be $\gamma=0.1$,
even though the performance was mostly insensitive to this parameter (see the supplement for details).

\textbf{Dynamic environments.}
The described training data generation procedure is limited to static environments, since the trajectory generation method is unable to take the changing geometry into account.
How can we use it to train a perception system that would be able to cope with dynamic environments?
Our key observation is that training on multiple static environments (for instance with varying gate positions) is sufficient to operate in dynamic environments at test time.
We collect data from a variety of layouts, generated by slightly moving the gates from their initial position.
We generate a global reference trajectory for each layout and train a network jointly on all of these.
This simple approach supports generalization to dynamic tracks, with the additional benefit of improving the robustness of the system.
%
%
%

\section{Experiments in Simulation}
\label{sec:experiments}
We perform an extensive evaluation, both in simulation and on a physical system and compare our approach to a set of state-of-the-art baselines (\camready{see supplement for details}).
We first present experiments in a controlled, simulated environment.
The aim of these experiments is to get a sense of the capabilities of the presented approach, and compare to a direct end-to-end deep learning approach that regresses body rates based on image data.
We use RotorS~\cite{Furrer_2016} and Gazebo for all simulation experiments.
\begin{figure}
\centering
\begin{subfigure}{.5\textwidth}
  \centering
  \includegraphics[width=.85\linewidth]{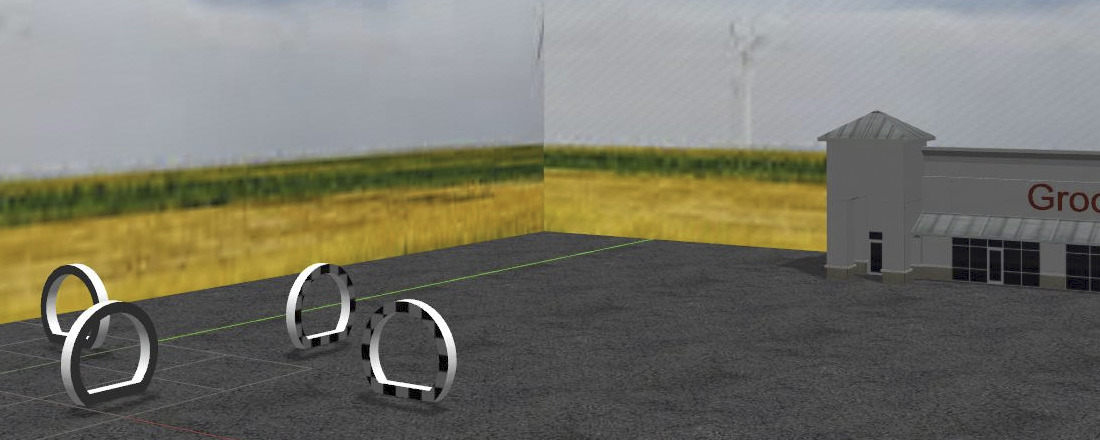}
  \caption{}
  \label{img:sim_track_small}
\end{subfigure}%
\begin{subfigure}{.5\textwidth}
  \centering
  \includegraphics[width=.85\linewidth]{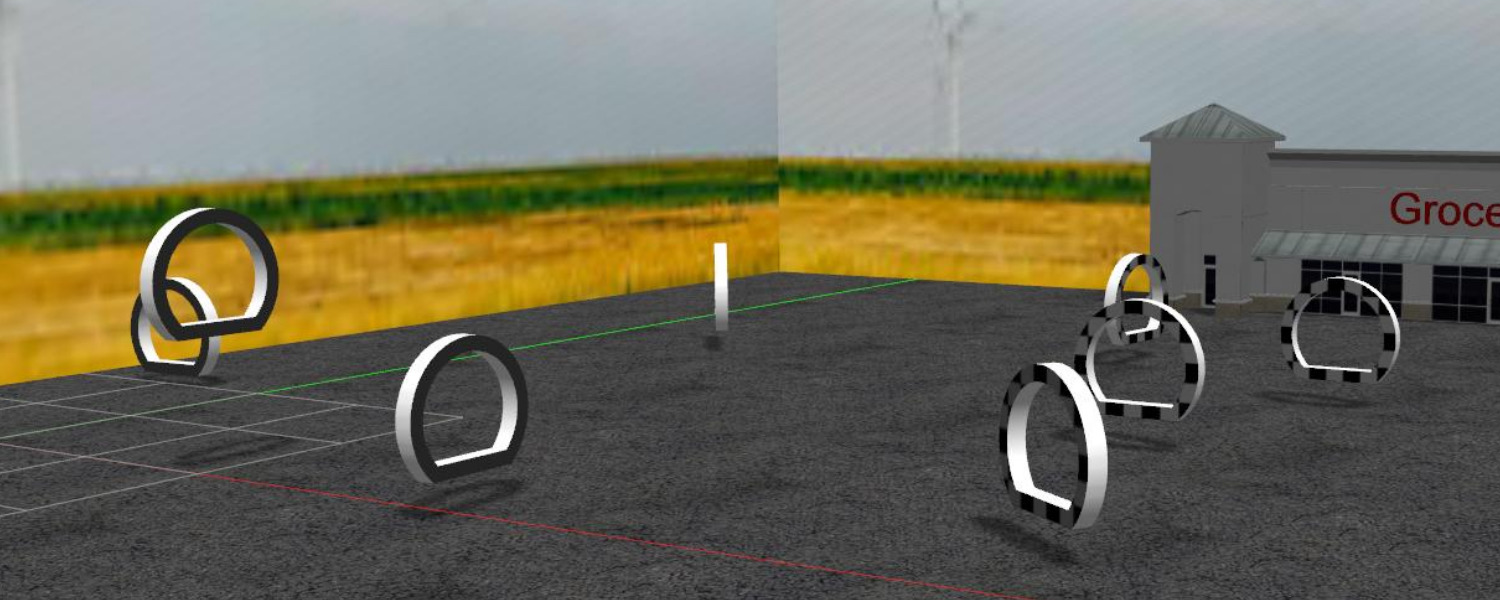}
  \caption{}
  \label{img:sim_track_large}
\end{subfigure}
\caption{\small Illustration of the small (\textbf{a}) and large (\textbf{b}) simulated tracks. The small track consists of 4 gates placed in a planar configuration and spans a total length of 43 meters. The large track consists of 8 gates placed at different heights and spans a total length of 116 meters.}
\vspace{-6mm}
\end{figure}

\subsection{Comparison to end-to-end learning approach}
In our first scenario, we use a small track that consists of four gates in a planar
configuration with a total length of 43 meters (Figure~\ref{img:sim_track_small}).
\camready{We use this track to compare the performance to a deep
learning baseline that directly regresses body rates from raw images~\cite{mueller2017teaching}.}
Ground truth body rates for the baseline were provided
by generating a minimum snap reference trajectory through all gates and then tracking it with a low-level controller~\cite{Faessler15icra}.
%

%
While our approach was always able to successfully complete the track, the naive baseline
could never pass through more than one gate.
Training on more data (35K samples, as compared to 5K samples used by our method) did not noticeably improve the performance of the baseline.
In contrast to previous work~\cite{mueller2017teaching}, we believe that end-to-end learning of low-level controls is suboptimal for the task of drone navigation when operating in the real world.
Indeed, the network has to learn the basic notion of stability from scratch in order to control an unstable platform such as a quadrotor~\cite{Narendra_1990}.
This leads to high sample complexity, and gives no mathematical guarantees on the platforms stability.
Additionally, the network is constrained by computation time. In order to guarantee stable control, the baseline network would have to produce control commands at a higher frequency than the camera images arrive and process them at a rate that is computationally infeasible with existing onboard hardware.
In contrast, our approach can benefit from years of study in the field of control theory~\cite{Hoffmann_2007}.
Because stability is handled by a classic controller, the network
can focus on the main task of robust navigation, which leads to high sample efficiency.
Additionally, because the network does not need to ensure the stability of the platform,
it can process images at a lower rate than the low-level controller,
which enables onboard computation.

Given its inability to complete even this simple track, we do not conduct any further experiments with the direct end-to-end regression baseline.

\subsection{Performance on a complex track}\label{sec:sim_experiment_static_track}
In order to explore the capabilities of our approach of performing high-speed
racing, we conduct a second set of experiments on a larger and more complex track (Figure~\ref{img:sim_track_large}) with 8 gates and a length of 116 meters.
%
%
The quantitative evaluation is conducted in terms of average task completion rate over five runs initialized with different random seeds.
For one run, the task completion metric linearly increases with each passed gate while 100\% task completion is achieved if the quadrotor is able to successfully complete five consecutive laps without crashing.
As baseline, we use a pure feedforward setting by following the global trajectory $t_g$ using visual inertial odometry~\cite{svo}.

\begin{figure}
\centering
\begin{subfigure}{.33\textwidth}
  \centering
  \includegraphics[width=1.05\linewidth]{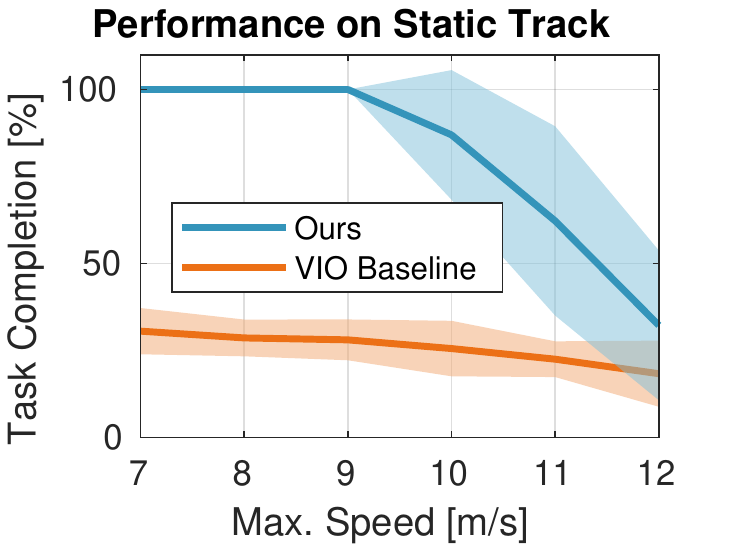}
  \caption{}
  \label{fig:comparison_max_speed}
\end{subfigure}%
\begin{subfigure}{.33\textwidth}
  \centering
  \includegraphics[width=1.05\linewidth]{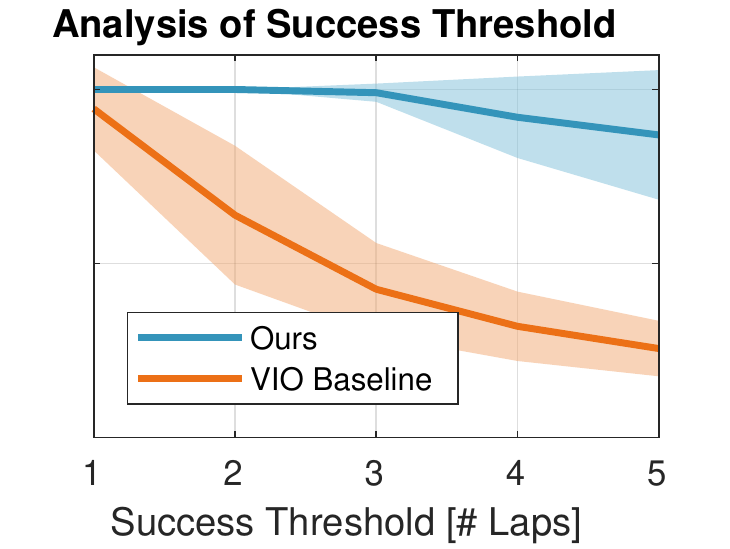}
  \caption{}
  \label{fig:comparison_success_threshold}
\end{subfigure}%
\begin{subfigure}{.33\textwidth}
  \centering
  \includegraphics[width=1.05\linewidth]{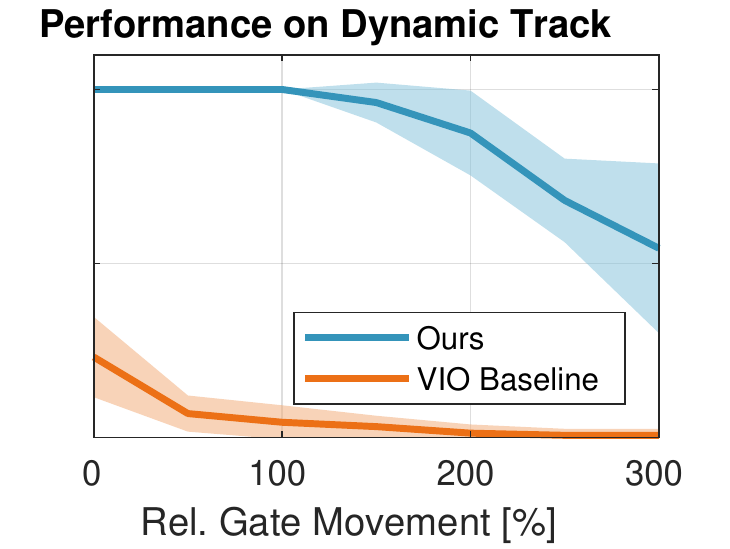}
  \caption{}
  \label{fig:comparison_moving_gates}
\end{subfigure}
\caption{\small
\textbf{a)} Results of simulation experiments on the large track with static gates for different maximum speeds.
\textit{Task completion rate} measures the fraction of gates that were successfully completed without crashing.
For each speed 10 runs were performed.
A task completion rate of $100\%$ is achieved if the drone can complete five consecutive laps without crashing.
\textbf{b)} Analysis of the influence of the choice of success threshold. The experimental setting is the same as in Figure~\ref{fig:comparison_max_speed}, but the performance is reported for a fixed maximum speed of $\SI{10}{\meter\per\second}$ and different success thresholds.
The $y$-axis is shared with Figure~\ref{fig:comparison_max_speed}.
\textbf{c)} Result of our approach when flying through a simulated track with moving gates.
Every gate independently moves with a sinusoidal pattern whose amplitude is equal to its base size (\SI{1.3}{\meter}) times the indicated multiplier.
For each amplitude 10 runs were performed.
As for the static gate experiment, a task completion rate of $100\%$ is achieved if the drone can complete five consecutive laps without crashing.
The $y$-axis is shared with Figure~\ref{fig:comparison_max_speed}.
The reader is encouraged to watch the supplementary video to better understand the experimental setup and task difficulty.}
\end{figure}

The results of this experiment are shown in Figure~\ref{fig:comparison_max_speed}.
We can observe that the VIO baseline performs inferior compared to our approach, on both the static and dynamic track.
On the static track, the VIO baseline fails due to the accumulated drift, as shown in Figure~\ref{fig:comparison_success_threshold}.
While the VIO baseline performs comparably when one single lap is considered a success, the performance degrades rapidly if the threshold for success is raised to more laps.
Our approach reliably works up to a maximum speed of \SI{9}{\meter\per\second}, while the performance gracefully degrades at higher velocities.
The decrease in performance at higher speeds is mainly due to the higher body rates of the quadrotor that larger velocities inevitably entail.
Since the predictions of the network are in the body frame, the limited prediction frequency (\SI{30}Hz in the simulation experiments) is no longer sufficient to cope with the large roll and pitch rates of the platform at high velocities.
%
%

%
%
%
\subsection{Generalization to dynamic environments}
The learned policy has a characteristic that the expert policy lacks: coping with dynamic environments. In those, waypoints and collision-free trajectories cannot be defined \emph{a priori}.
To quantitatively test this ability, we reuse the track layout from the previous experiment (Figure~\ref{img:sim_track_large}), but dynamically move each gate according to a sinusoidal pattern in each dimension independently.
Figure~\ref{fig:comparison_moving_gates} compares our system to the VIO baseline for varying amplitudes of gates' movement relative to their base size.
We evaluate the performance using the same metric as explained in Section~\ref{sec:sim_experiment_static_track}.
For this experiment, we kept the maximum platform velocity $v_{max}$ constant at \SI{8}{\meter\per\second}.
Despite the high speed, our approach can handle dynamic gate movements of up to 1.5 times the gates' diameter without crashing.
In contrast, the VIO baseline (\emph{i.e.} the expert policy) cannot adapt to changes in the environment, and fails even for tiny gate motions.
The performance of our approach gracefully degrades for gate movements larger than 1.5 times the gates' diameter, mainly due to the fact that consecutive gates get too close in flight direction while being shifted in other directions.
Such configurations require extremely sharp turns that go beyond the navigation capabilities of the system.
%
From this experiment, we can conclude that our approach reactively adapts to dynamic changes and generalizes well to cases where the track remains roughly similar to the one collected data from.

\begin{minipage}{\textwidth}
  \begin{minipage}[b]{0.4\textwidth}
    \centering
    \includegraphics[width=0.9\linewidth]{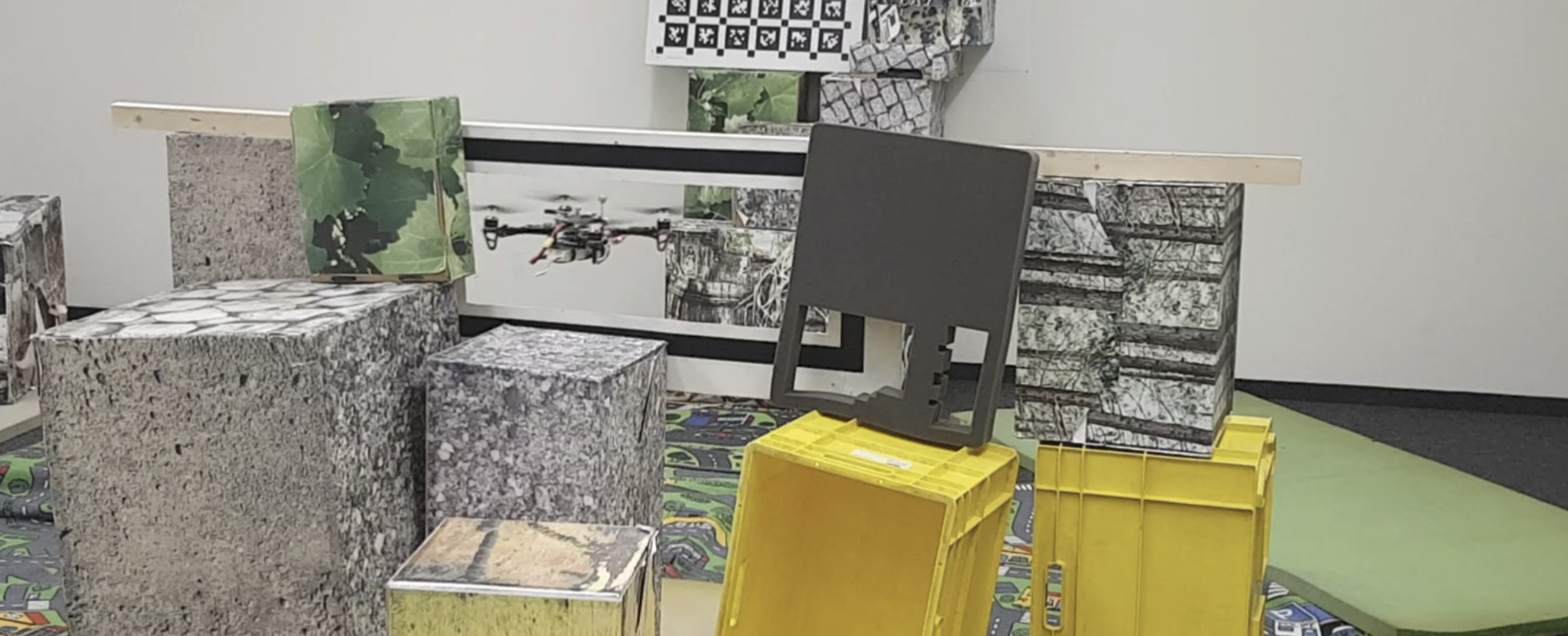}
    \captionof{figure}{ \small Setup of the narrow gap and occlusion experiments.}
\label{img:narrow_gap}
  \end{minipage}
  \hfill
  \begin{minipage}[b]{0.59\textwidth}
    \centering
    \small
	\begin{tabular}{c | c c}
		\toprule
		Relative Angle Range [$\degree$] & Handcrafted Detector & Network \\
		\midrule
		$[0, 30]$ & $70\%$ & $100\%$ \\
		$[30, 70]$ & $0\%$ & $80\%$ \\
		$[70, 90]$* & $0\%$ & $20\%$ \\
		\bottomrule
	\end{tabular}
      \captionof{table}{\small Success rate for flying through a narrow gap from different initial angles. Each row reports the average of ten runs uniformly spanning the range. The gate was completely invisible at initialization in the experiments marked with *.}
	\label{tab:static_window}
    \end{minipage}
  \end{minipage}

\section{Experiments in the Physical World}

To show the ability of our approach to control real quadrotors, we performed experiments on a physical platform.
%
%
We compare our approach to state-of-the-art classic
approaches to robot navigation, as well as to human drone pilots of different skill levels.
For these experiments, we collected data in the real world.
Technical details on the platform used can be found in the supplement.

In a first set of experiments the quadrotor was required to pass through a  narrow gate, only slightly larger than the platform itself.
These experiments are designed to test the robustness and precision of the proposed approach.
%
An illustration of the setup is shown in Figure~\ref{img:narrow_gap}.
We compare our approach to the handcrafted window detector of Falanga et al.~\cite{aggressive_falanga} by replacing our perception system (Section~\ref{sec:perception_system}) with the handcrafted detector and leaving the control system (Section~\ref{sec:control_system}) unchanged.

Table~\ref{tab:static_window} shows a comparison between our approach and the baseline. We test the robustness of both approaches to the initial position of the quadrotor.
To do so we place the platform at different starting angles with respect to the gate (measured as the angle between the line joining the center of gravity of the quadrotor and the gate, respectively, and the optical axis of the forward facing camera on the platform).
We measure average success rate to pass the gate without crashing.
The experiments indicate that our approach is robust to initial conditions. The drone is able to pass the gate consistently, even if the gate is only partially visible.
By contrast, the handcrafted baseline cannot detect the gate if it's not entirely in the field of view.
The baseline sometimes fails even if the gate is fully visible because the window detector loses tracking due to platform vibrations.

\begin{figure}[t]
\centering
\begin{subfigure}{.5\textwidth}
  \centering
  \includegraphics[width=1.0\linewidth]{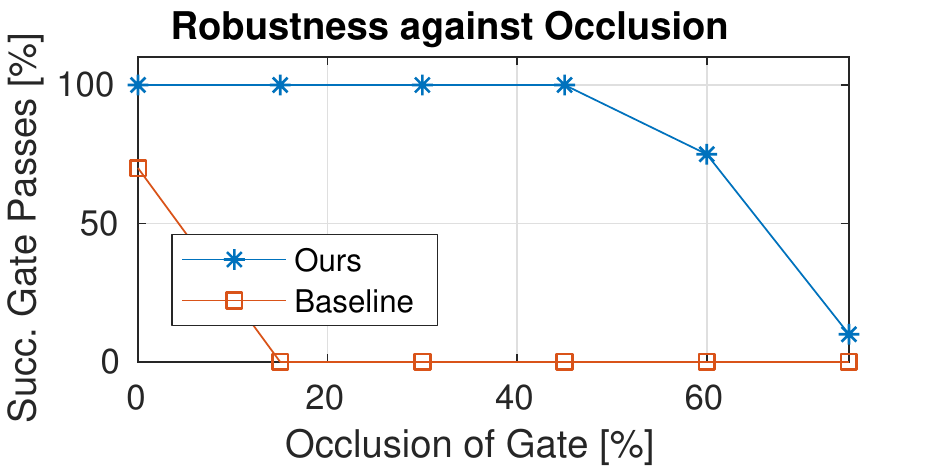}
  \caption{}
  \label{fig:comparison_occlusion}
\end{subfigure}%
\begin{subfigure}{.5\textwidth}
  \centering
  \includegraphics[width=1.0\linewidth]{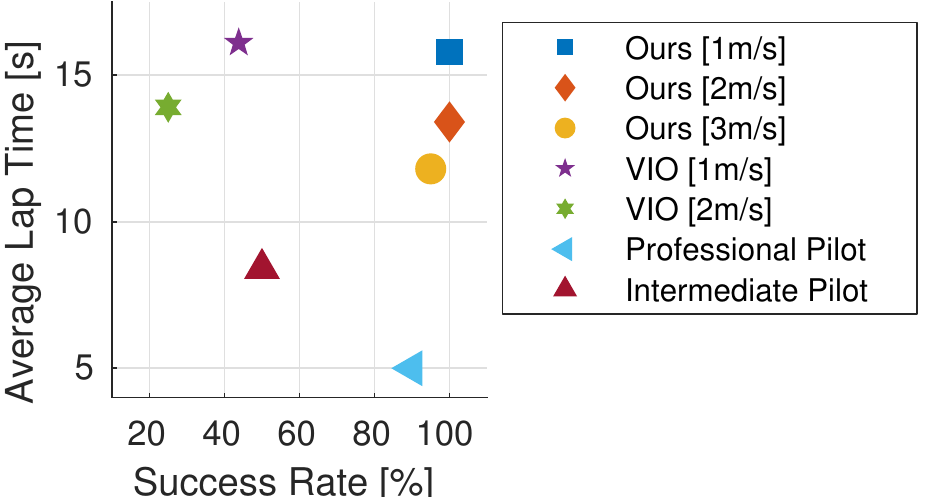}
  \caption{}
  \label{fig:comparison_real_world}
\end{subfigure}
\caption{\small \textbf{a)} Success rate for different amount of occlusion of the gate.
The area is calculated on the entire size of the gate.
At more than $60\%$ occlusion, the platform has barely any space to pass through the gap.
\textbf{b)} Results on a real race track composed of 4 gates.
Our learning-based approach compares favorably against a set of baselines based on visual-inertial state estimation.
Additionally, we compare against an intermediate and a professional drone pilot.
We evaluate success rate using the same metric as explained in Section~\ref{sec:sim_experiment_static_track}.
}
\vspace{-5mm}

\end{figure}

In order to further highlight the robustness and generalization abilities of the approach, we perform experiments
with an increasing amount of clutter that occludes the gate.
Note that the learning approach has never seen these configurations during training.
Figure~\ref{fig:comparison_occlusion} shows that our approach is robust to occlusions
of up to 50\% of the total area of the gate (Figure~\ref{img:narrow_gap}), whereas the handcrafted baseline
breaks down even for moderate levels of occlusion.
For occlusions larger than 50\% we observe a rapid drop in performance. This can be explained by the fact that the remaining gap was barely larger than the drone itself, requiring very high precision to successfully pass it. Furthermore, visual ambiguities of the gate itself become problematic. If just one of the edges of the window is visible, it is impossible to differentiate between the top and bottom part. This results in over-correction when the drone is very close to the gate.

%
%

\subsection{Experiments on a race track}\label{sec:exp_real_track}
In the last set of experiments we challenge the system to race through a track with either static or dynamics gates.
%
The track is shown in Figure~\ref{fig:catch_eye_1}. It is composed of four gates
and has a total length of 21~meters.
%
To fully understand the potential and limitations of our approach we compared to a diverse set of baselines, such as a classic approach based on planning and tracking~\cite{LoiBruMcGKum17}
and human pilots of different skill levels.
Note that due to the smaller size of the real track compared to the simulated one, the maximum speed achieved in real world experiments is lower than in simulation.
For our baseline, we use a state-of-the-art visual-inertial
odometry approach~\cite{LoiBruMcGKum17} to provide global state estimates
in order to track the global reference trajectory.
%
%
%

Figure~\ref{fig:comparison_real_world} summarizes the quantitative results of our evaluation,
where we measure success rate (completing five consecutive laps without crashing),
as well as the best lap time.
Our learning-based approach outperforms the visual odometry-based baseline, whose drift at high speeds inevitably leads to poor performance.
By generating waypoint commands in body frame, instead, our approach is insensitive to state estimation drift, and can complete the track
with higher robustness and speed than the VIO baseline.

In order to see how state-of-the-art autonomous approaches compare to human pilots, we asked a professional
and an intermediate pilot to race through the track in first-person view. We allowed the pilots
to practice the track for 10 laps before lap times and failures were measured.
It is evident from Figure~\ref{fig:comparison_real_world} that both the professional
and the intermediate pilots were able to complete the track faster than the
autonomous systems. The high speed and aggressive flight by human pilots comes at
the cost of increased failure rates, however. The intermediate pilot
in particular had issues with the sharp turns present in the track, leading to frequent
crashes.
Compared with the autonomous systems, human pilots perform more agile maneuvers, especially in sharp turns.
Such maneuvers require a level of reasoning about the environment that our autonomous system still lacks.

In a last qualitative experiment, we manually moved gates while the quadrotor navigated through the track. This requires the navigation system
to be able to reactively respond to dynamic changes.
Note that moving gates break the main assumption of traditional high-speed navigation approaches~\cite{mit_plan, teach_repeat},
specifically that the trajectory can be pre-planned in a static world. They could thus not be deployed in this scenario.
Due to the dynamic nature of this experiment, we encourage the reader to watch the supplementary video\footnote{Available from: \url{http://youtu.be/8RILnqPxo1s}}.
As in the simulation experiments, the system can generalize to dynamically moving gates on the real track.
It is worth noting that training data was collected by changing the position of only a single gate, but the network is able to cope with movement of any gate at test time.

\section{Discussion}
\label{sec:conclusions}
We have presented a new approach to autonomous, vision-based drone
racing. Our method uses a compact convolutional neural network to continuously predict
a desired waypoint and a desired speed directly from raw images. These high-level commands are then executed
by a classic control stack. To enable agile and fast flight, we train the network to follow
a global reference trajectory. The system combines the robust perceptual awareness
of modern machine learning pipelines with the stability and speed
of well-known control algorithms.

We demonstrated the capabilities of this
integrated approach to perception and control in an extensive set of experiments
on real drones and in simulation. Our experiments show that the resulting system is
able to robustly navigate complex race tracks, avoids the problem of drift that
is inherent in systems relying on global state estimates, and can cope with
highly dynamic and cluttered environments.

While our current set of experiments was conducted in the context of drone racing,
we believe that the presented approach could have broader implications for building
robust robot navigation systems that need to be able to act in a highly dynamic
world. Methods based on geometric mapping, localization and planning have inherent
limitations in this setting. Hybrid systems that incorporate machine learning,
like the one presented in this paper, can offer a compelling solution to this task,
given the possibility to benefit from near-optimal solutions to
different subproblems.

Scaling such hybrid approaches to more general environments \camready{and tasks} is an exciting avenue for future work
that poses several challenges.
First, while the ability of our system to navigate through moving or partially occluded gates is promising, performance will degrade if the appearance of the environment changes substantially beyond what was observed during training.
Second, in order to train the perception system, our current approach requires a significant amount of data in the application environment.
This might be acceptable in some scenarios, but not practical when fast adaptation to previously unseen environments is needed.
This could be addressed with techniques such as few-shot learning.
Third, in the cases where trajectory optimization cannot provide a policy to be imitated, for instance in the presence of extremely tight turns, the learner is also likely to fail.
This issue could be alleviated by integrating learning deeper into the control system.




\acknowledgments{This work was supported by the Intel Network on Intelligent Systems, the Swiss National Center of Competence Research Robotics (NCCR), through the Swiss National Science Foundation, and the SNSF-ERC starting grant.}


{\small
\bibliography{bibliography/references}  
}

\clearpage
\section*{Supplementary Material}

\subsection{Trajectory generation}
\textbf{Generation of global trajectory.}
Both in simulation and in real world experiments, a global trajectory is used to generate ground truth labels. 
To generate the trajectory, we use the implementation from Mellinger et al. \cite{mellinger2011minimum}.
The trajectory is generated by providing a set of waypoints to pass through, a maximum velocity to achieve as well as constraints on maximum thrust and body rates.
Note that the speed on the global trajectory is not constant. 
As waypoints, the centers of the gates are used. 
Furthermore, the trajectory can be shaped by additional waypoints, for example if it would pass close to a wall otherwise. 
In both simulation and real world experiments, the maximum normalized thrust along the trajectory was set to $\SI{18}{\meter\per\square\second}$ and the maximum roll and pitch rate to $\SI{1.5}{\radian\per\second}$.
The maximum speed was chosen based on the dimensions of the track.
For the large simulated track, a maximum speed of $\SI{10}{\meter\per\second}$ was chosen, while on the smaller real world track $\SI{6}{\meter\per\second}$.

\textbf{Generation of trajectory segments.}
The proposed navigation approach relies on constant re-computation of trajectory segments $t_s$ based on the output of a CNN. 
Implemented as state-interception trajectories, $t_s$ can be computed by specifying a start state and a goal state.
While the start state of the trajectory segment is fully defined by the quadrotor's current position, velocity, and acceleration, the end state is only constrained by the goal position $p_g$, leaving velocity and acceleration in that state unconstrained.
This is, however, not an issue, since only the first part of each trajectory segment is executed in a receding horizon fashion.
Indeed, any time a new network prediction is available, a new state interception trajectory $t_s$ is calculated.

The goal position $p_g$ is dependent on the prediction horizon $d$ (see Section~\ref{sec:training_procedure}), which directly influences the aggressiveness of a maneuver.
A long prediction horizon leads to a smoother flight pattern, usually required on straight and fast parts of the track.
Conversely, a short horizon performs more agile maneuvers, usually required in tight turns and in the proximity of gates.

The generation of the goal position $p_g$ differs from training time to test time. 
At training time, the quadrotors current position is projected onto the global trajectory and propagated by a prediction horizon $d_{train}$. 
At test time, the output of the network is back-projected along the camera projection ray by a planning length $d_{test}$. 
%

At training time, we define the prediction horizon $d_{train}$ as a function of distance from the next gate:
\begin{equation}
d_{train} = \max \left(d_{min}, \min \left(\Vert s_{last} \Vert , \Vert s_{next} \Vert\right)\right) \, ,
\label{eq:d_train}
\end{equation}
where $s_{last} \in \mathbb{R}^3$ and $s_{next} \in \mathbb{R}^3$ are the distances from the last gate and the next gate to be traversed, respectively,
and~$d_{min}$ represents the minimum prediction horizon.
In our simulated track experiment, a minimum prediction horizon of $d_{min} = \SI{1.5}{\meter}$ was used, while for the real world track $d_{min} = \SI{1.0}{\meter}$ was used.

At test time, since the output of the network is a direction and a velocity, the length of a trajectory segment needs to be computed. 
To distinguish the length of trajectory segments at test time from the same concept at training time, we call it at test time \textit{planning length}. 
The planning length of trajectory segments is computed based on the velocity output of the network (computation based on the location of the quadrotor with respect to the gates is not possible at test time since we do not have knowledge about gate positions).
The objective is again to adapt the planning length such that both smooth flight at high speed and aggressive manoeuvres in tight turns are possible. 
We achieve this versatility by computing the planning length according to this linear function:
\begin{equation}
d_{test} = \min \left[d_{max}, \max \left(d_{min}, m_d v_{out}\right)\right] \, ,
\label{eq:d_test}
\end{equation}
where $m_d = \SI{0.6}{\second}$, $d_{min}=\SI{1.0}{\meter}$ and $d_{max}=\SI{2.0}{\meter}$ in our real world experiments and $m_d = \SI{0.5}{\second}$, $d_{min}=\SI{2.0}{\meter}$ and $d_{max}=\SI{5.0}{\meter}$ in the simulated track experiment. 

\subsection{Training data generation}

In this section, the generation of training data in both simulation and real world experiments is explained in detail. While in simulation, data is generated while flying, in real world experiments we collect data by carrying the quad through the track. Both approaches will be explained in the following sections. 

\textbf{Generate data in simulation.}
In our simulation experiment, we perform a modified version of DAgger~\cite{dagger} to train our flying policy.
On the data collected through the expert policy (Section~\ref{sec:training_procedure}) (in our case we let the expert fly for $\SI{40}{\second}$),
the network is trained for 10 epochs on the accumulated data. 
In the following run, the trained network is predicting actions, which are only executed if they keep the quadrotor within a margin $\epsilon$ from the global trajectory. 
In case the network's action violates this constraint, the expert policy is executed, generating a new training sample. 
This procedure is an automated form of \cite{dagger}, and allows the network to recover when deviating from the global trajectory. 
After another $\SI{40}{\second}$ of data generation, the network is retrained on all the accumulated data for 10 epochs. 
As soon as the network performs well on a given margin $\epsilon$, the margin is increased. 
This process repeats until the network can eventually complete the whole track without help of the expert policy. 
In our simulation experiments, the margin $\epsilon$ was set to $\SI{0.5}{\meter}$ after the first training iteration.
The margin was incremented by $\SI{0.5}{\meter}$ as soon as the network could complete the track with limited help from the expert policy (less than 50 expert actions needed). 
For the experiment on the static track, 20k images were collected, while for the dynamic experiment 100k images of random gate positions were generated.

\textbf{Generate data in real world.}
In contrast to simulated experiments, in real world we do not want the network to fly until it is fully trained.
Instead of using the DAgger approach, we ensure sufficient coverage of the possible actions by carrying the quadrotor through the track.
During this procedure which we call \textit{handheld} mode, the expert policy is constantly generating training samples. 
Since we do not risk crashing the quad into obstacles, it is easy to collect training samples from potentially dangerous positions like close to gates or other obstacles. 
Due to the drift of onboard state estimation, data is generated for a small part of the track before the quadrotor is reinitialized at a known position.
For the experiment on the static track, 25k images were collected, while for the dynamic experiment an additional 15k images were collected for different gate positions. 
For the narrow gap and occlusion experiments, 23k images were collected.

\subsection{Cross-Validation of loss weight $\gamma$}
\begin{figure}
\centering
\includegraphics[width=.9\linewidth]{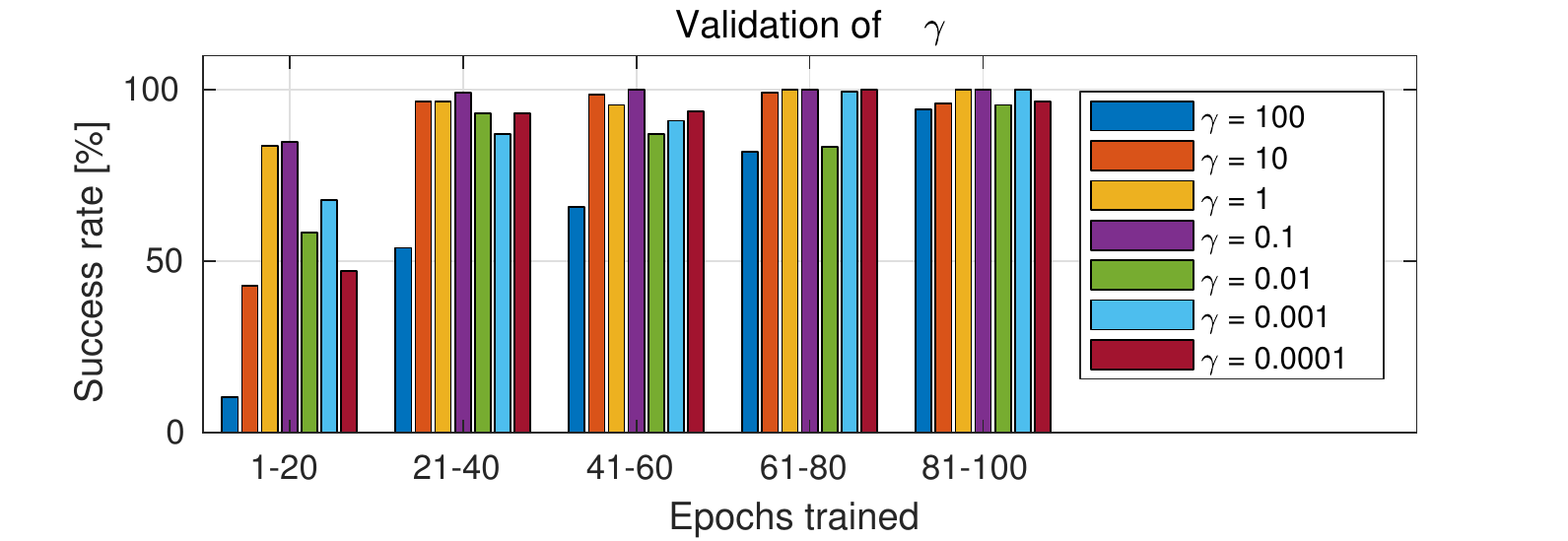}
\caption{Success rate for different values of $\gamma$.
For each $\gamma$, the network is trained up to 100 epochs. 
Performance is evaluated after each training epoch according to the performance criterion defined in~\ref{sec:sim_experiment_static_track}.
For readability reasons, performance measurements are averaged over 20 epochs. 
}
\label{fig:gamma_validation}
\end{figure}

The loss function used to train our network, defined in Eq.~\ref{eq:loss},
depends on the weighting factor $\gamma$ for the velocity MSE.
We performed a cross-validation of $\gamma$ to understand its influence on the system performance as a function of training time.
Specifically, we selected 7 values of $\gamma$ in the range $[0.0001,100]$ equispaced in logarithmic scale.
Our network was then trained for up to 100 epochs on data generated from the static simulated track (Figure~\ref{img:sim_track_large}). 
After each epoch, performance is tested at a speed of $\SI{8}{\meter\per\second}$ according to the performance measure defined in~\ref{sec:sim_experiment_static_track}.
Figure~\ref{fig:gamma_validation} shows the results of this evaluation.
The network shows to be able to complete the track for all configurations after 80 epochs.
Despite some values for $\gamma$ lead to faster learning, we see that the system performance is not extremely sensitive to this weighting factor.
Since $\gamma=0.1$ proved to give the best results, we used it in all our further experiments.
 
\subsection{Baselines}
\camready{
There are three baselines against which we compared our system.
The first baseline is an end-to-end low level commands' (body-rates) regressor. Ground truth for this baseline is generated by the low-level controller in~\cite{faessler2016autonomous} while the drone is tracking the global reference trajectory under the assumption of accurate state estimation. Then, a CNN is used to generate those commands from images only.  Images are recorded at 30Hz by the on-board camera and the latest low-level controls produced by the controller are associated to them. To increase the diversity of data, and ease the shift between learner and teacher, we perform the positional DAGGER procedure explained in the supplement in Section 6.2. Nonetheless, this baseline was unable to complete even the simplest track layout.
The second baseline is reported in Section 4.3, where we show that our system significantly outperforms a handcrafted gate detector even for a very simple gate configuration (a black square on a white background). In this baseline, the relative orientation between the gate and the drone is generated by the handcrafted detector described in~\cite{aggressive_falanga} instead of a CNN. The control structure remains unchanged. 
Finally, our strongest baseline is the standard approach for drone navigation: track a global trajectory with noisy state estimates obtained by a visual-inertial localization system. Specifically, it consists of executing the expert policy (Sec. 3.1) with noisy state estimates.
This is referred to as “VIO baseline”.
}

\subsection{Platform}
In all the real world track experiments, we deployed an in-house quadrotor equipped with an Intel UpBoard and a Qualcomm Snapdragon Flight Kit.
While the latter is used for visual-inertial odometry\footnote{Available from: \url{https://developer.qualcomm.com/software/machine-vision-sdk}}, the former represents the main computational unit of the platform.
%
%
The Intel UpBoard was used to run all the calculations required for flying, from neural network prediction to trajectory generation and tracking.
This hardware configuration allowed for a network inference rate of approximately $\SI{10}{\hertz}$.

\subsection{Network architecture and Grad-CAM}

We implement the perception system by a convolutional network.
The input to the network is a $300\times200$ pixel RGB image, captured from the onboard camera at a frame rate of $\SI{30}{\hertz}$.
After normalization in the $[0,1]$ range, the input is passed through 7 convolutional layers, divided in 3 residual blocks, and a final fully connected layer that
outputs a tuple~$\lbrace \vec{x}, v \rbrace$.
$\vec{x} \in [-1,1]^2$ is a two-dimensional vector that encodes the direction to the new goal in normalized image coordinates and $v \in [0,1]$ is a normalized desired speed to approach it.

\begin{figure*}[t]
\centering
\def\colwidth{0.23\columnwidth}
\addtolength{\tabcolsep}{-4pt}
\begin{tabular}{m{\colwidth} m{\colwidth} m{\colwidth} m{\colwidth}}
\centering
  \includegraphics[width=\linewidth]{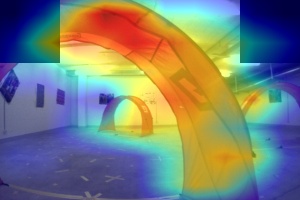} &
  \includegraphics[width=\linewidth]{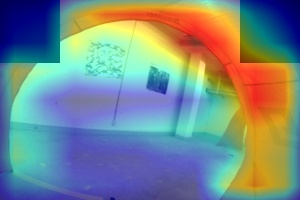} &
  \includegraphics[width=\linewidth]{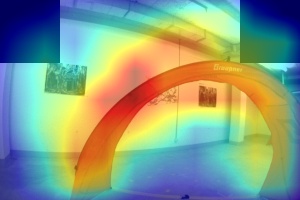} &
  \includegraphics[width=\linewidth]{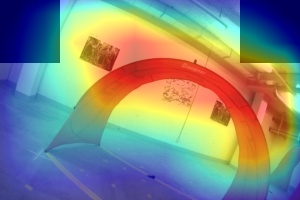} \\
\end{tabular}
\addtolength{\tabcolsep}{4pt}
\vspace{-2ex}
\captionof{figure}{Visualization of network attention using the Grad-CAM technique \cite{gradcam}. Yellow to red
areas correspond to areas of medium to high attention, while blue corresponds to areas of low attention.
It is evident that the network learns to mostly focus on gates instead of relying on the background, which explains
its capability to robustly handle dynamically moving gates. (Best viewed in color.)}
\label{fig:gradcams}
\end{figure*}

To understand why the network is robust to previously unseen changes in the environment, we visualize the networks attention using the Grad-CAM \cite{gradcam} technique in Figure~\ref{fig:gradcams}. 
Grad-CAM visualizes which parts of an input image were important for the decisions made by the network.
It becomes evident that the network bases its decision mostly on the visual input that is
most relevant to the task at hand --- the gates --- while it mostly ignores the background.

\end{document}